\documentclass{article}

\usepackage{PRIMEarxiv}
\usepackage{amsmath}
\usepackage[utf8]{inputenc} 
\usepackage[T1]{fontenc}    
\usepackage{hyperref}       
\usepackage{url}            
\usepackage{booktabs}       
\usepackage{amsfonts}       
\usepackage{nicefrac}       
\usepackage{microtype}      
\usepackage{lipsum}
\usepackage{fancyhdr}       
\usepackage{graphicx}       
\graphicspath{{media/}}     

\title{Dance Recalibration for dance coherency with Recurrent Convolution Block
}

\author{
  Seungho Eum, Ihjoon Cho, Junghyeon Kim \\
  Computer Science \\
  Sogang Univ. \\
  Seoul, Korea\\
  \texttt{\{seunghoeum, dlwns23\}@sogang.ac.kr} \\
  \texttt{wjdgus3006@naver.com}
}
\begin{document}
\maketitle

\begin{abstract}
  With the recent advancements in generative AI such as GAN, Diffusion, and VAE, the use of generative AI for dance generation has seen significant progress and received considerable interest. In this study, We propose R-Lodge, an enhanced version of Lodge. R-Lodge incorporates  Recurrent Sequential Representation Learning named Dance Recalibration to original coarse-to-fine long dance generation model. R-Lodge utilizes Dance Recalibration method using \(N\) Dance Recalibration Block to address the lack of consistency in the coarse dance representation of the Lodge model. By utilizing this method, each generated dance motion incorporates a bit of information from the previous dance motions. We evaluate R-Lodge on FineDance dataset and the results show that R-Lodge enhances the consistency of the whole generated dance motions.
\end{abstract}


\section{Introduction}

Recently, due to the increase in data and improvement in computing power, generative AI technology has experienced rapid growth, leading to significant advancements in dance generation AI as well. Generating novel dances from effective generative model not only can enhance the diversity in visual arts but also reduce dance creating costs and make the process more efficient. Game animators, movie directors, and etc. can save time and help create more immersive and realistic animations.

Previous works \cite{kim2022brand2, ofli2011learn2dance, siyao2022bailando, zhou2019dance} can generate short dances which is not proper in actual applications that requires dances longer than a few seconds. However, actual applications generally require dances lasting 3 to 5 minutes, and some performances even require dances lasting several hours. Therefore, there is now a demand for the generation of longer dance sequences.To generate long dance, dealing with the high computational costs is needed. Various methods such as auto-regressive model \cite{huang2020dance, li2021ai, zhuang2022music2dance, valle2021transflower}  try to deal with it but they are failed due to their lack of mode coverage. 

Building on the exceptional generation performance of diffusion models, \cite{tseng2023edge} and \cite{li2024lodge} propose music-conditioned dance generation models based on diffusion. \cite{tseng2023edge} introduced a transformer-based diffusion network for long sequences dance generation. They are able to generate arbitrarily long sequences by chaining the shorter clips with local consistency but lack an extreme-long-term dependencies. Additionally, \cite{tseng2023edge} proposed a novel Contact Consistency Loss and Physical Foot Contact Score to eliminate foot sliding physical implausibilities but still such problems are observed in generated frames.

\cite{li2024lodge} attempts to address Edge's challenge regarding the consistency of combined long dance sequences. \cite{li2024lodge} propose a coarse-to-fine two stage diffusion framework and characteristic dance primitives to produce long dances in a parallel manner. The first stage gets long music input to generate dance primitives which is coarse-grained diffusion. And the parallel local diffusion modules follow to generate short dance segments finally concatenated into long dance sequences. This two-stage coarse-to-fine diffusion framework strikes a between overall choreographic patterns and the quality of short-term local movements. \cite{li2024lodge} also proposes a method to overcome the foot-sliding problem, termed the foot refine block to eliminate artifacts.


Although Lodge could quickly generate long dance sequences, the process of generating coarse movements led to instability and a lack of consistency with previous frames, resulting in awkward transitions in the dance movements. To address this, we added a recurrent recalibration process during the coarse dance generation stage to enhance consistency. This addition ensures that subsequent dance movements are smooth and natural. Our contributions can be summarized as follows:
\newline
\begin{itemize}
    \item A Recurrent Block was added to the Coarse Dance Decoder.
    \item The Recurrent Block enhances the consistency of the dance and ensures that the movements are not awkward.
    \item Achieved state-of-the-art performance in dance coherency on the FineDance \cite{li2023finedance} dataset.
\end{itemize}

\section{Related Works}
\subsection{Human motion generation}
Human motion generation is one of the most interested field in computer vision studies. Due to the high level of interest, it has witnessed remarkable advances through many researches. Despite this significant growth, human motion generation remains a challenging problem, often plagued by issues such as foot-sliding and a lack of consistency or smoothness due to abrupt motion changes.

Roughly human motion generation is based on either regression models or generative models. In recent years, generative models have become the dominant approach. For instance, Kinetic-GAN \cite{degardin2022generative} utilizes a GAN combined with GCN. It leverages latent space disentanglement to separate different aspects of motion, facilitating easier manipulation and diverse generation. Furthermore, stochastic variations help produce a wider range of realistic motions.

MDM \cite{tevet2022human} and MLD \cite{chen2023executing}, for example, utilize diffusion model to generate desired motions. \cite{ahn2018text2action, ahuja2019language2pose, ghosh2021synthesis, petrovich2022temos, athanasiou2022teach} are among the studies to generate human motion sequences from text descriptions. Since above methods can face challenges in zero-shot generation, \cite{tevet2022motionclip, hong2022avatarclip} employs CLIP, a pre-trained vision-language model, for better zero-shot generation performance.

VQ-VAE is another popular method for generating human motion. For instance, T2MT \cite{guo2022tm2t} uses VQ-VAE for training text-to-motion and motion-to-text tasks. T2M-GPT \cite{zhang2023t2m} leverages a GPT-like transformer architecture and VQ-VAE combined with an EMA. 

\subsection{Music To Dance Generation}
Dance and music are inseparable that music provides the foundation for the movement and emotion of dance while we express the characteristic of music by dance. Accordingly, abundant studies make a goal to generate quality dance conditioned on music.

The generation of dance sequences driven by music has been an active area of research, focusing on synchronizing dance movements with musical inputs. Traditional motion-graph methods approached this as a similarity-based retrieval problem, limiting the diversity and creativity of the generated dance sequences. However, recent advancements in deep learning have led to more aesthetically appealing results.

Sequence-based methods using LSTM and Transformer networks predict subsequent dance frames in an autoregressive manner. For example, FACT \cite{li2021ai} inputs music and seed motions into a Transformer network to generate new dance frames frame by frame. However, challenges such as error accumulation and motion freezing persist. VQ-VAE is another well-known approach, as seen in methods like Bailando \cite{siyao2022bailando}, which incorporates reinforcement learning to optimize rhythm and maintain high motion quality. However, the pre-trained codebook in VQ-VAE can limit diversity and hinder generalization.

GAN-based methods like MNET \cite{kim2022brand} employ adversarial training to produce realistic dance clips and achieve genre control, though they often face issues like mode collapse and training instability. Diffusion-based methods have shown significant progress in generating high-quality dance clips. For instance, EDGE \cite{tseng2023edge} uses diffusion inpainting to generate consistent dance segments, while FineDance \cite{li2023finedance} introduces diffusion models to produce diverse and high-quality dance sequences.

Lodge \cite{li2024lodge} represents a notable advancement in this domain, employing a coarse-to-fine diffusion framework to generate extremely long dance sequences. By introducing characteristic dance primitives, Lodge ensures both global choreographic patterns and local motion quality. Additionally, the foot refine block in Lodge addresses artifacts such as foot-skating, enhancing the physical realism of the generated dances. However, Lodge has its limitations. It currently cannot generate dance movements with hand gestures or facial expressions, which are crucial elements in dance performances. Moreover, while Lodge aims to produce diverse dance sequences, its reliance on characteristic dance primitives may limit the overall diversity of the generated dances. This limitation opens avenues for future research to further enhance both the realism and the variety of the dance sequences.
\subsection{Generation for Sequential Data}
The processing of sequential data based on deep learning initially started with Recurrent Neural Networks \cite{schmidt2019recurrent}. \cite{schmidt2019recurrent} showed good performance at the time by utilizing information from previous sequences. However, \cite{schmidt2019recurrent} had the gradient vanishing problem. To overcome this limitation, models such as \cite{sak2014long} and \cite{dey2017gate} were introduced, which use gates to enable learning the amount of information from previous sequences. Although these models are complex, they are considered good research contributions as they solve the gradient vanishing problem.

Additionally, with the advancement of Transformers, the performance in processing sequential data has improved further. Particularly in vision-related tasks involving sequential data, such as \cite{ge2022long, yan2021videogpt, yu2023magvit, hong2022cogvideo, ma2024latte} and \cite{arnab2021vivit, liu2022video, fan2021multiscale,li2022mvitv2}, incorporating Transformers has yielded good performance. However, Transformers require high computational costs and have significant computational demands. Although methods like \cite{ahmad2021fpn, lee2023afi} are used to alleviate these issues, they still demand expensive computational resources.

We discovered that in the Lodge model, which generates coarse dances using \cite{ho2020denoising, song2020denoising}, the generated dances do not sequentially connect. To address this issue, we conducted research to utilize the dance information from previous frames in order to create coarse dance information that includes sequential frame data.

%

\section{Methodology}
We applied Recurrency Sequence Processing to address the lack of consistency in the coarse dance representation of the~\cite{li2024lodge} model. We named this Recurrency Sequence Representation Learning as Dance Recalibration (DR). Dance recalibration uses \(n\) Dance Recalibration Blocks (DRB) corresponding to the length of the rough dance sequence to add sequential information to the rough dance representation to improve the consistency of the whole dance. The overall structure of our model is illustrated in Figure 1.

\begin{figure}[!t]
    \centering
    \includegraphics[width=\textwidth]{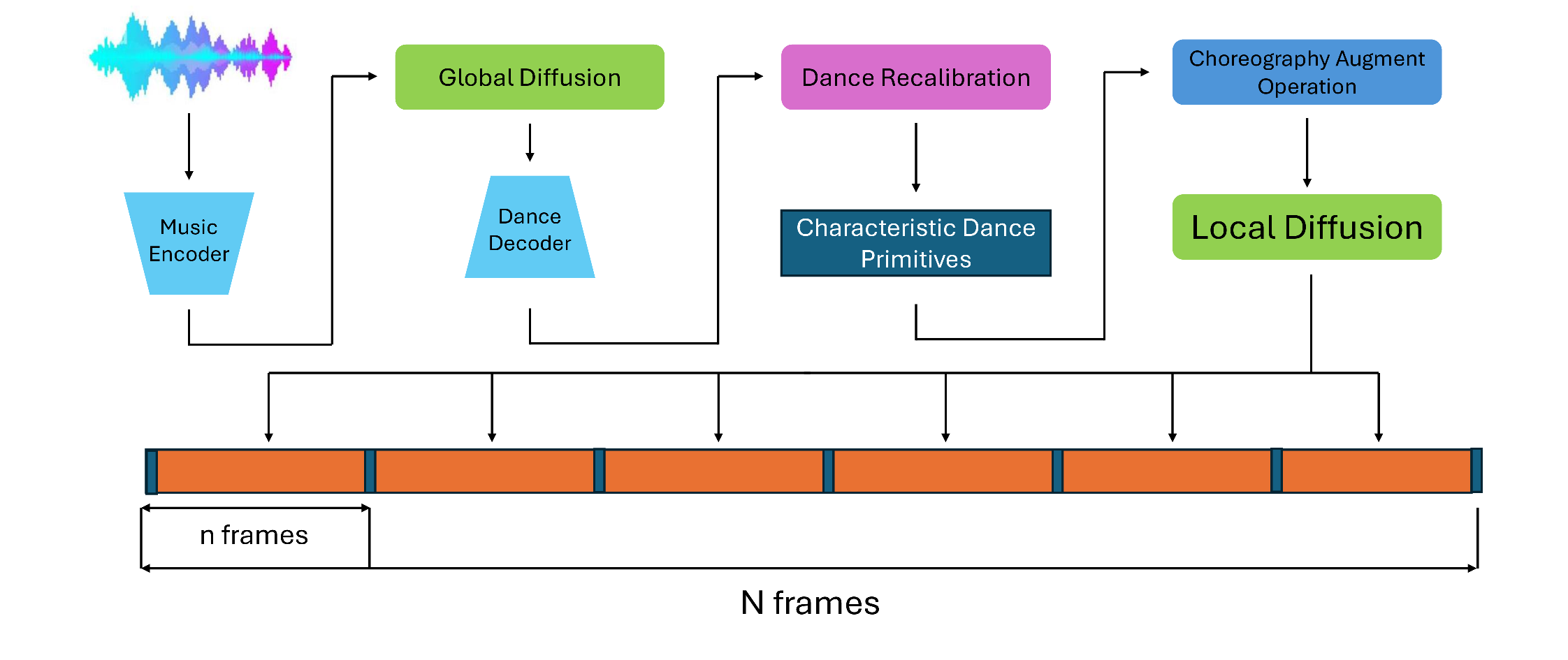}
    \caption{overall procedure of Pooling processing by our Pooling Block}
    \label{fig:enter-label4}
\end{figure}

\subsection{Dance Recalibration (DR)}
When the dance motion representation passes through the Dance Decoder Process using the~\cite{li2024lodge} model, it yields a coarse dance motion representation. During this process, the dance motion representations that pass through Global Diffusion follow a distribution but can output unstable values. This results in awkward dance motions when viewed from a sequential perspective. To address this issue, we added a Dance Recalibration Process.

DR fundamentally follows a structure similar to RNNs. Although RNNs are known to suffer from the gradient vanishing problem as they get deeper, the sequence length of the coarse dance representation in \cite{li2024lodge} is not long enough to cause this issue, making it suitable for use. Using LSTM or GRU, which solve the gradient vanishing problem, would make the model too complex and computationally expensive, making them unsuitable for use with the Denoising Diffusion Process \cite{ho2020denoising, song2020denoising}.

The coarse dance representation has 139 channels, consisting of 4-dim foot positions, 3-dim root translation, 6-dim rotaion information and 126-dim joint rotation channels. Of these, the 126-dim channels directly impact the dance motion, and all DR operations are performed using these 126 channels.

The values output from the Global Dance Decoder \(GD_{i}\), contain unstable dance motion information that follows a distribution. We construct Global Recalibrated Dance \(GRD_{i}\) by concatenating \(C\) the information from \(GRD_{i-1}\) with \(GD_{i}\) and applying pooling \(P\), thereby adding sequential information. However, using previous information as is may result in overly simple and smoothly connected dance motions. To prevent this, we add Gaussian noise \(G\) to the previous information \(GRD_{i-1}\) to produce more varied dance motions. This process is represented in Equations 1 below. The entire procedure is illustrated in Figure 2, 3.
\begin{equation}
    GRD_{i} = P(C(GD_{i} , GRD_{i-1} + G(Threshold))
\end{equation}

\begin{figure}[!t]
    \centering
    \includegraphics[width=\textwidth]{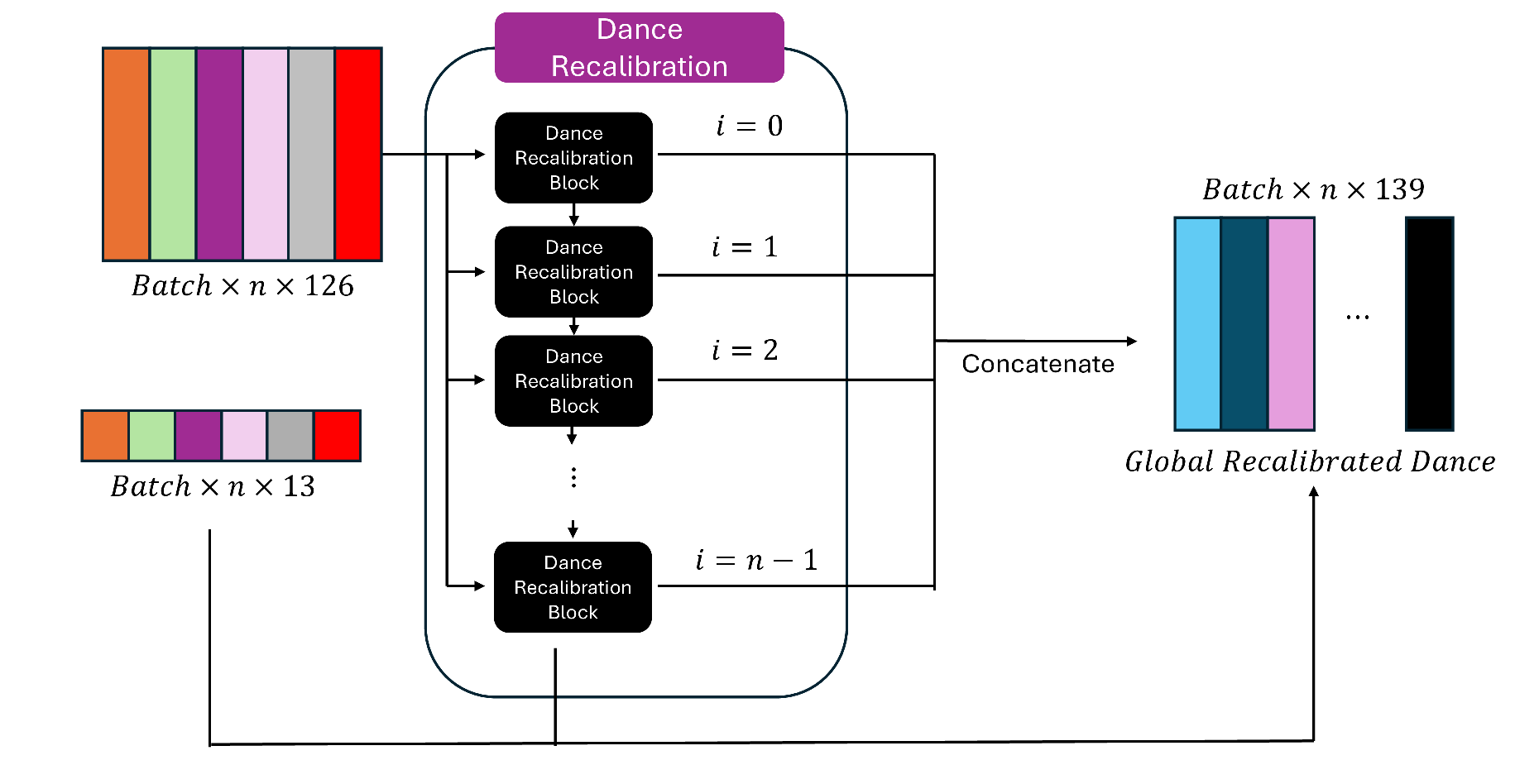}
    \caption{Overall of the Dance Recalibration Block Structure}
    \label{fig:enter-label1}
\end{figure}

\begin{figure}[!t]
    \centering
    \includegraphics[width=\textwidth]{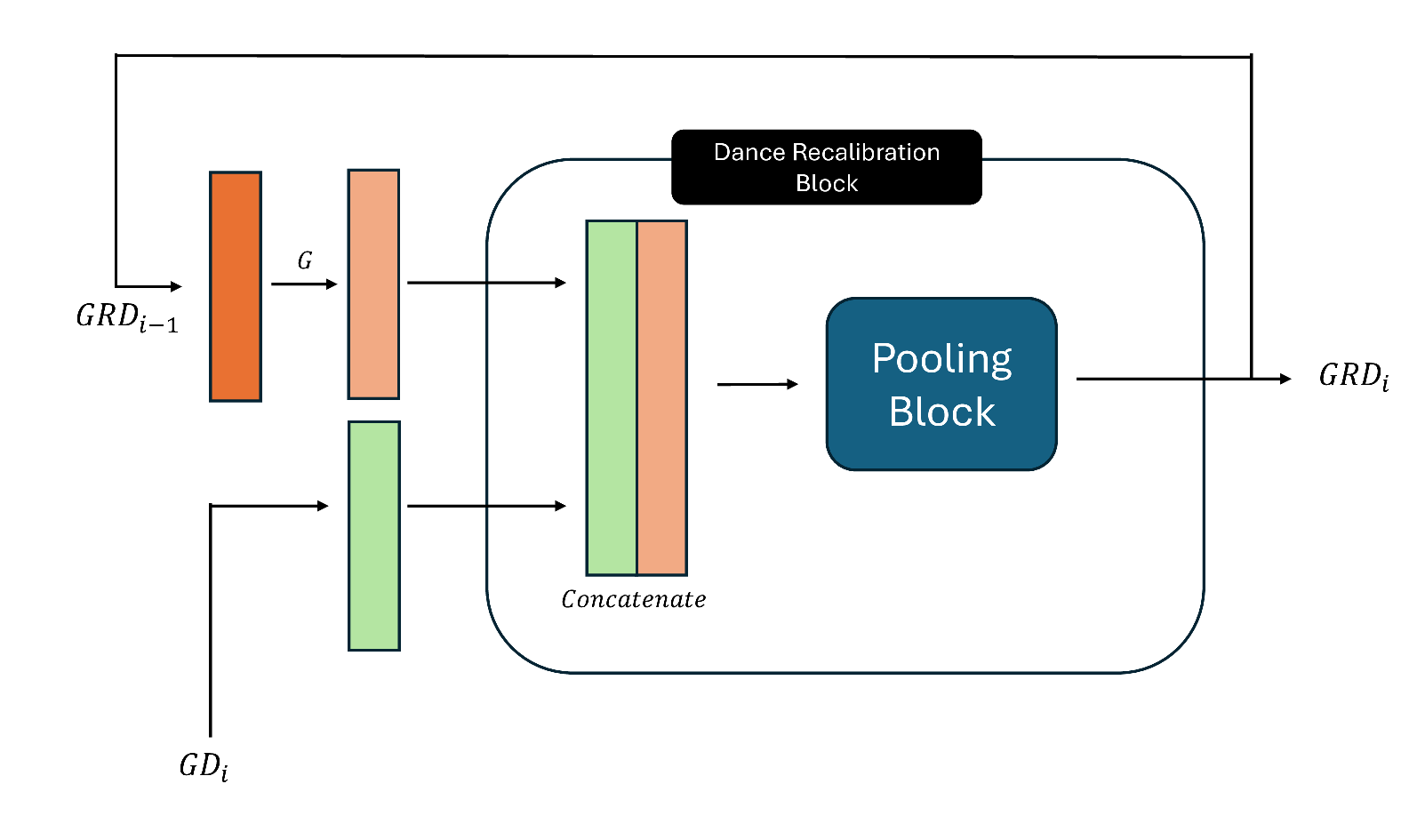}
    \caption{The structure of the dance recalibration block}
    \label{fig:enter-label2}
\end{figure}

\subsection{Pooling Block}
Pooling \(P\) uses a simple pooling method. When \(GRD_{i}\) with added \(G\) and \(GD_{i+1}\) are input, they are concatenated into a \((Batch\times2\times126)\). First, we perform Layer Normalization to minimize differences between layers. Then, we pass through three simple 1D-Convolution Blocks, each followed by an activation function and batch normalization, to construct \(GRD_{i+1}\) that includes information from the previous dance motion. This procedure is illustrated in Figure 4.

\begin{figure}[!t]
    \centering
    \includegraphics[width=\textwidth]{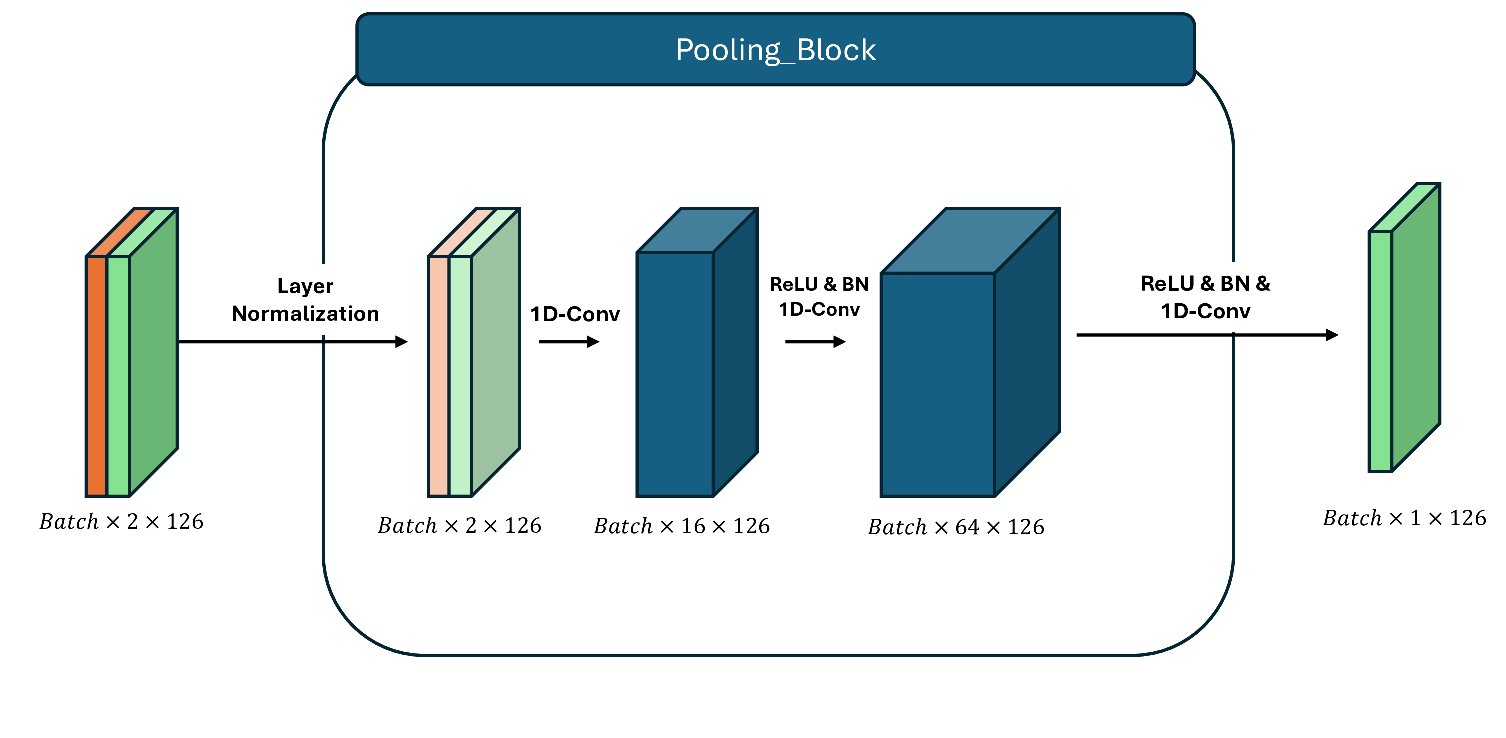}
    \caption{overall procedure of Pooling processing by our Pooling Block}
    \label{fig:enter-label3}
\end{figure}

By following all these steps, each dance motion incorporates a bit of information from the previous dance motions, producing an overall coarse dance motion that follows the distribution of Global Diffusion while also retaining sequential information. This process is expressed in Equation 2:

\begin{equation}
    Total Coarse Dance Motion = C_{i=1}^{n}(P(C(GD_{i} , GRD_{i-1} + G(Threshold))), P(GD_{0}))
\end{equation}

We did not use bi-directional information because it complicates the calculations and can destabilize sequential information when using more than two \(GD_{i}\). Since there is a trade-off between generating complex dance motions and maintaining consistency, it is crucial to add appropriate noise. However, due to time constraints, we could not conduct various ablation studies.

\section {Experiments}

\paragraph{Datasets} We use public music-dance paired dataset for music-driven dance generation, FineDance \cite{li2023finedance}, which includes 22 dance genres. FineDance claims to contain 14.6 hours of music-dance paired data with detailed hand motions and accurate postures. However, it actually has 7.7 hours, totaling 831,600 frames at a frame rate of 30fps, and includes 16 different dance genres. Our entire model is trained and tested on this dataset. We create dance sequences, each comprising 1024 frames long dance sequences (equivalent to 34.13 seconds), using music from the test set of the FineDance dataset as input.
\paragraph{Implementation details} 
We follow the experimental settings of Lodge for the global music feature, the local music feature and the generation of characteristic dance primitives. The global music feature extracted from the music and audio analysis library Librosa \cite{mcfee2015librosa} has a length of 1024 (34.13 seconds), and the local music feature input for the Local Diffusion has a length of 256 (8.53 seconds). As output of the global diffusion, we obtain 13 characteristic dance primitives which consisting of 5 coarse dance motions and 8 fine dance motions. These 8 fine dance motions are then used as input for choreography augmentation augment to 16 fine dance motions by mirroring and aligned with the input music's beat.

\subsection{Comparison on the FineDance dataset}
In this section, we compare our method with the several past works including Lodge. FACT \cite{li2021ai} and MNET \cite{kim2022brand}are auto-regressive dance generation methods. Bailando \cite{siyao2022bailando} is a follow-up approach that employs VQ-VAE to transform dance movements into tokens and achieve outstanding qualitative performance. EDGE represents a significant advancement in the field of dance generation by utilizing diffusion models to achieve substantial performance improvements. Specifically, EDGE introduces a transformer-based diffusion model paired with Jukebox, conferring powerful editing capabilities and setting a new vision in generating realistic and physically plausible dance motions. Lodge is a network designed to generate extremely long dance sequences conditioned on music. It employs a two-stage coarse-to-fine diffusion architecture with characteristic dance primitives as intermediate representations between two diffusion models. It significantly outperforms existing models in generating coherent, high-quality, and expressive dance sequences.

\paragraph{Motion Quality \& Diversity}
Motion quality is assessed using two primary measures. The Frechet Inception Distance (FID) measures the distance between the generated dance motion features and the ground truth dance sequences, indicating the quality of the motion. Separate FID scores are reported for kinematic features ($\mathrm{FID}_k$) and geometric features ($\mathrm{FID}_g$). Additionally, the Foot Skating Ratio (FSR) calculates the proportion of frames where the feet slide on the ground instead of making solid contact, with lower values indicating better physical realism. Motion diversity is evaluated through the diversity in kinematic features (Div$_k$), which assesses the variety in the generated dance motions based on kinematic properties such as speed and acceleration. The diversity in geometric features (Div$_g$) evaluates the diversity of generated dance movements by examining geometric properties and predefined movement templates.

\paragraph{Beat Aligment Score (BAS)}
The Beat Alignment Score (BAS) measures how well the generated dance sequences align with the beats of the accompanying music, reflecting the synchronization between music and dance.

\paragraph{Production Efficiency}
Production efficiency is measured by the average run time required to generate a specific length of dance sequence, such as 1024 frames, to evaluate the computational efficiency of the model. All experiments were conducted on the same computer equipped with 8 Nvidia A100 GPUs.

\begin{table}[ht]
    \centering
    \caption{Comparison with state-of-the-art methods on the FineDance dataset. Wins represent the ratio of victories Lodge (DDPM) achieved in the user study.}
    \resizebox{\textwidth}{!}{
    \begin{tabular}{@{\hskip 3pt}l@{\hskip 3pt}c@{\hskip 3pt}c@{\hskip 3pt}c@{\hskip 3pt}c@{\hskip 3pt}c@{\hskip 3pt}c@{\hskip 3pt}c@{\hskip 3pt}c}
        \toprule
        Method & \text{FID$_k$} $\downarrow$ & \text{FID$_g$} $\downarrow$ & \text{FSR} $\downarrow$ & \text{Div$_k$} $\uparrow$ & \text{Div$_g$} $\uparrow$ & \text{BAS} $\uparrow$ & \text{Run Time} $\downarrow$\\
        \midrule
        Ground Truth & / & / & 6.22\% & 9.73 & 7.44 & 0.2120 &\\
        \midrule
        FACT \cite{li2021ai} & 113.38 & 97.05 & 28.44\% & 3.36 & 6.37 & 0.1831 & 35.88s \\
        MNET \cite{kim2022brand} & 104.71 & 90.31 & 39.36\% & 3.12 & 6.14 & 0.1864 & 38.91s \\
        Bailando \cite{siyao2022bailando} & 82.81 & 28.17 & 18.76\% & 7.74 & 6.25 & 0.2029 & 5.46s  \\
        EDGE \cite{tseng2023edge} & 94.34 & 50.38 & 20.04\% & 8.13 & 6.45 & 0.2116 & 8.59s\\
        Lodge (DDIM) \cite{song2020denoising} & 50.00 & 35.52 & 2.76\% & 5.67 & 4.96 & 0.2269 & 4.57s \\
        Lodge (DDPM) \cite{ho2020denoising} & 45.56 & 34.29 & 5.01\% & 6.75 & 5.64 & 0.2397 & 30.93s \\
        \midrule
        (Lodge + Ours) (DDPM) & \textbf{42.76} & \textbf{32.17} & 4.97\% & 5.79 & 5.24 & \textbf{0.2501} & 31.17s \\
        \bottomrule
    \end{tabular}}
    \label{tab:comparison}
\end{table}

As seen in Table 1, Our proposed model showed an approximately 4\% improvement in BAS scores compared to the baseline model Lodge, and the FID scores increased by an average of about 6\%, achieving state-of-the-art (SOTA) performance. Of course, to fully validate our proposal, experiments on various benchmarks are necessary. However, it is significant that we improved performance on the widely-used FineDataset without significantly increasing computer resources or runtime.
\section{Conclusion and Future Work}
\paragraph{Conclusion} In this work, we propose R-Lodge enhanced version of Lodge which incorporates Dance Recalibration Using Recurrency Block. Proposed method address the burst changes between adjacent coarse dance motions from original Lodge which interfere the smoothness in generated dance motion. Through the evaluation of our generated samples, R-Lodge demonstrates smoother dance motions than Lodge while maintaining a comparable level of overall dance coherence. However, our experiment is limited to the FineDance dataset\cite{li2023finedance}, which hinders generalization to other datasets. Furthermore, we have not yet proposed a solution to improve the weak motion diversity, which is a limitation of the original Lodge approach.
\paragraph{Future Work} For future work, as we mentioned earlier, experimentation with other datasets, e.g. AIST++\cite{li2021ai}, is necessary to demonstrate that our approach is not limited to FineDance\cite{li2023finedance} dataset. Following this, our next objective is to increase the diversity of dance motions generated from input music via music frame genre embedding. Original Lodge reflects the common music genre embedding for whole frames before Local Diffusion. We aim to extract the genre of each individual music frame and integrate it with the corresponding frame of music feature. This integration will enable the global diffusion process to be conditioned not only on the music frame feature but also on its genre embedding. We anticipate it will not compromise the coherence of overall dance motions but rather increase diversity by embodying the subtle variations in genre features from each frame of the music.


\bibliographystyle{plain}
\bibliography{references}

\end{document}